\documentclass[10pt,twocolumn,letterpaper]{article}

\usepackage{wacv}
\usepackage{times}
\usepackage{epsfig}
\usepackage{graphicx}
\usepackage{amsmath}
\usepackage{amssymb}
\usepackage{booktabs}
\usepackage{tabulary}
\usepackage{multirow}

\def\wacvPaperID{351} 

\wacvalgorithmstrack   

\wacvfinalcopy 

\ifwacvfinal
\else
\fi
\usepackage[pagebackref=true,breaklinks=true,colorlinks,bookmarks=false]{hyperref}
\begin{document}

\thispagestyle{empty}
\title{Zero-shot Object Detection Through Vision-Language Embedding Alignment}

\author{Johnathan Xie$\thanks{Work was done at Dawnlight Technologies Inc.}$\\
Stanford University\\
{\tt\small jwxie@stanford.edu}
\and
Shuai Zheng$^{*}$\\
{\tt\small kyle@kylezheng.com}
}
\maketitle
\begin{abstract}
Recent approaches have shown that training deep neural networks directly on large-scale image-text pair collections enables zero-shot transfer on various recognition tasks. One central issue is how this can be generalized to object detection, which involves the non-semantic task of localization as well as semantic task of classification. To solve this problem, we introduce a vision-language embedding alignment method that transfers the generalization capabilities of a pretrained model such as CLIP to an object detector like YOLOv5. We formulate a loss function that allows us to align the image and text embeddings from the pretrained model CLIP with the modified semantic prediction head from the detector. With this method, we are able to train an object detector that achieves state-of-the-art performance on the COCO, ILSVRC, and Visual Genome zero-shot detection benchmarks. During inference, our model can be adapted to detect any number of object classes without additional training. We also find that standard object detection scaling can transfer well to our method and find consistent improvements across various scales of YOLOv5 models and the YOLOv3 model. Lastly, we develop a self-labeling method that provides a significant score improvement without needing extra images nor labels.
\end{abstract}
\section{Introduction}
\label{sec:intro}
Zero-shot detection (ZSD)~\cite{Bansal_eccv_2018} aims to train a model to detect unseen objects. Despite its difficulty, zero-shot object detection has many applications. For example, to develop a Chimpanzee detector~\cite{Schofield19}, animal scientists needed to collect specialized training data and exhaustively label for Chimpanzees, but with a generally trained zero-shot object detector they could simply detect using new reference embeddings. 

Typical object detectors are trained on data with a fixed number of categories. The PASCAL VOC dataset~\cite{Everingham_2010} has 20 classes, while the COCO dataset~\cite{Tsung-yi-eccv-2014} has 80 classes. These datasets have inspired many important object detection methods such as DPM~\cite{lsvm-pami}, R-CNN~\cite{Girshick_cvpr2014,girshick_iccv2015,Ren_nips_2015}, SSD~\cite{liu_eccv_2016}, and YOLO~\cite{Redmon_cvpr_2016,Redmon_cvpr_2017,glenn_jocher_2022_6222936}.

While research in detecting rare categories with higher accuracy has shown extensive progression, models continue to score considerably lower for rare objects~\cite{gupta2019lvis,Tan_2020_LVIS_1st}. Furthermore, gathering detection data for a large number of categories is difficult due to the long-tailed nature of objects and because exhaustively labeling many class types is expensive. In contrast, gathering text image pairs is a far easier task as these images can be sourced from the internet to assemble massive training datasets such as YFCC100M~\cite{kalkowski_mm2015}. Using this dataset, Radford~\etal~\cite{Radford_clip} demonstrated impressive zero-shot capabilities across a wide array of computer vision datasets with varying category granularity such as~\cite{berg_birdsnap_cvpr2014,Jia_deng_cvpr_2009,stanford_cars,FGVC_aircraft}, but they have not explored the zero-shot capabilities for object detection. In this paper we devise a method to transfer this zero-shot capability to object detection.

To this end, we propose a method for adapting a one-stage detector to perform the ZSD task through aligning detector semantic outputs to embeddings from a trained vision-language model. Compared to previous ZSD works that only learn from text embedding alignment, such as~\cite{Bansal_eccv_2018,Rahman_accv_2018,li_aaai_2019,wang_ictai_2020,zhao_aaai_2020,zhu2020don,zheng_accv_2020,Zheng_zsis_cvpr2021}, we also consider image embedding alignment in our proposed method. We find that this addition of image embeddings to our proposed loss function provides a significant improvement to model performance. To obtain image embeddings, we crop ground truth bounding boxes and feed them through the CLIP~\cite{Radford_clip} image encoder. Model class outputs are then aligned using an $\mathcal{L}_1$ loss function. Furthermore, we create a new post-processing operation tailored to the ZSD task for YOLO~\cite{Redmon_cvpr_2016} style models to better detect unseen classes in scenes with both seen and unseen classes. Lastly, we also show that our method achieves consistent improvements on the different YOLOv5 network sizes. In summary, our contributions are:
\begin{itemize}
\item We show it is possible to generalize the capability of the pre-trained vision-language model CLIP~\cite{Radford_clip} to address zero-shot detection. Specifically, we train  ZSD-YOLO, our one stage zero-shot detection model~\cite{glenn_jocher_2022_6222936} that aligns detector semantic outputs to embeddings from a contrastively trained vision-language model CLIP~\cite{Radford_clip}.
\item We develop a self-labeling data augmentation method that provides 0.8 mAP and 1.7 mAP improvement on the 48/17~\cite{Bansal_eccv_2018} and 65/15~\cite{Rahman_aaai_2020} COCO ZSD splits respectively without needing extra images nor labels.
\item We propose a ZSD tailored post-processing function that provides a 1.6 mAP and 3.6 mAP improvement on the 48/17~\cite{Bansal_eccv_2018} and 65/15~\cite{Rahman_aaai_2020} COCO ZSD splits respectively.
\item We demonstrate that there are consistent improvements for the proposed ZSD-YOLO model with different network sizes which benefits model deployment under varying computational constraints.
\end{itemize}

\section{Related works}
\label{sec:related}
\paragraph{\textbf{Zero-shot learning.}}
Zero-shot learning (ZSL)~\cite{Xian_zsl_ce_pami2019} aims at developing a model that can identify data class types whose categories are not seen during training. In ZSL~\cite{Wang_zsl_survey_acm2019}, there are still some labeled training instances in the feature space and the categories of these instances are referred to as seen classes and the unlabeled instances are considered unseen classes. Prior works~\cite{Dietterich_Bakiri_jair_1995,Farhadi_cvpr_2009,Palatucci_nips_2009,Lampert_cvpr_2009} in ZSL focus on attribute learning which focuses on learning to recognize the properties of objects. However, these approaches treat all dimensions of the attribute space equally, which is sub-optimal because certain attributes are more important than others for a given task. More recently, CLIP~\cite{Radford_clip} uses contrastive learning that jointly trains an image encoder and a text encoder on a subset of~\cite{kalkowski_mm2015} to create a network which predicts the correct pairings of batched image-text training examples. During inference, the learned text encoder synthesizes a zero-shot linear classifier by embedding the names or descriptions of the target dataset's unseen classes. 

\paragraph{\textbf{Zero-shot detection.}}
Zero-shot detection (ZSD) was introduced through Bansal ~\etal~\cite{Bansal_eccv_2018} and describes the task of detecting objects that have no labeled samples in the training set. Bansal~\etal~\cite{Bansal_eccv_2018} established a baseline model through aligning model outputs to word-vector embeddings through linear projection and created the 48/17 benchmarking split based on the COCO~\cite{Lin_TY_COCO_2014} detection  dataset. Zhu~\etal~\cite{Zhu_tcst_2018} used the YOLOv1~\cite{Redmon_cvpr_2016} detector to improve zero-shot object recall. Li~\etal~\cite{li_aaai_2019} developed an attention mechanism to address zero-shot detection. Rahman~\cite{Rahman_aaai_2020} adopts the RetinaNet detector for zero-shot detection, using a polarity loss to increase the distance between class embeddings and introduced the 65/15 ZSD split on the COCO dataset. Zhao~\etal~\cite{zhao_aaai_2020} used a GAN to synthesize the semantic representation for unseen objects to help detect unseen objects. Zhu~\etal~\cite{Zhu_cvpr_2020} proposed DELO that synthesizes the visual features for unseen objects from semantic information and incorporates the synthesized features for the detection of seen and unseen classes Zheng~\etal~\cite{Zheng_zsis_cvpr2021} proposed a method to address the task of zero-shot instance segmentation using a method that consists of zero-shot detector, semantic mask head, background aware RPN, and synchronized background strategy. Hayat~\etal~\cite{hayat_accv_2020} requires knowledge of which classes will be used for unseen detection while training as well as their semantic embeddings, while ours and prior works such as~\cite{Bansal_eccv_2018,Rahman_accv_2018,Rahman_aaai_2020,Zheng_zsis_cvpr2021} can train without having to set unseen classes in training and can therefore run inference on a new set of unseen objects without any retraining. For this reason our paper as well as previous zero-shot detection papers such as~\cite{Zheng_zsis_cvpr2021} have not compared with their scores in benchmarking. 

\paragraph{Open Vocabulary Detection}
Very recently, Zareian~\etal~\cite{Zareian/cvpr2021} created a new dataset split of the COCO dataset based on the 2017 train and validation splits and introduced a task known as open-vocabulary detection (OVD). This new task does not eliminate instances of unseen classes in the detector training data. Gu~\etal~\cite{gu2021openvocabulary} has explored the idea of
improving open vocabulary detection, based on knowledge distillation and the two-stage Mask-RCNN~\cite{kaiming_he_iccv_2017} model.

In contrast to the prior works, our paper explores aligning one stage detector semantic outputs to CLIP embeddings to address the task of ZSD. To this end, we develop a new training algorithm for embedding alignment and modify the typical YOLOv5 model architecture to support ZSD. To optimize our method, we tailor the typical YOLOv5 postprocessing method to ZSD and develop a self-labeling data augmentation method to expand the knowledge of a zero-shot detector without additional data. We also explore model scaling, another crucial aspect of traditional detection systems that allows a model architecture to fit various applications with different computational constraints.

\begin{figure*}[tbp]
\centering
   \includegraphics[width=0.8\linewidth]{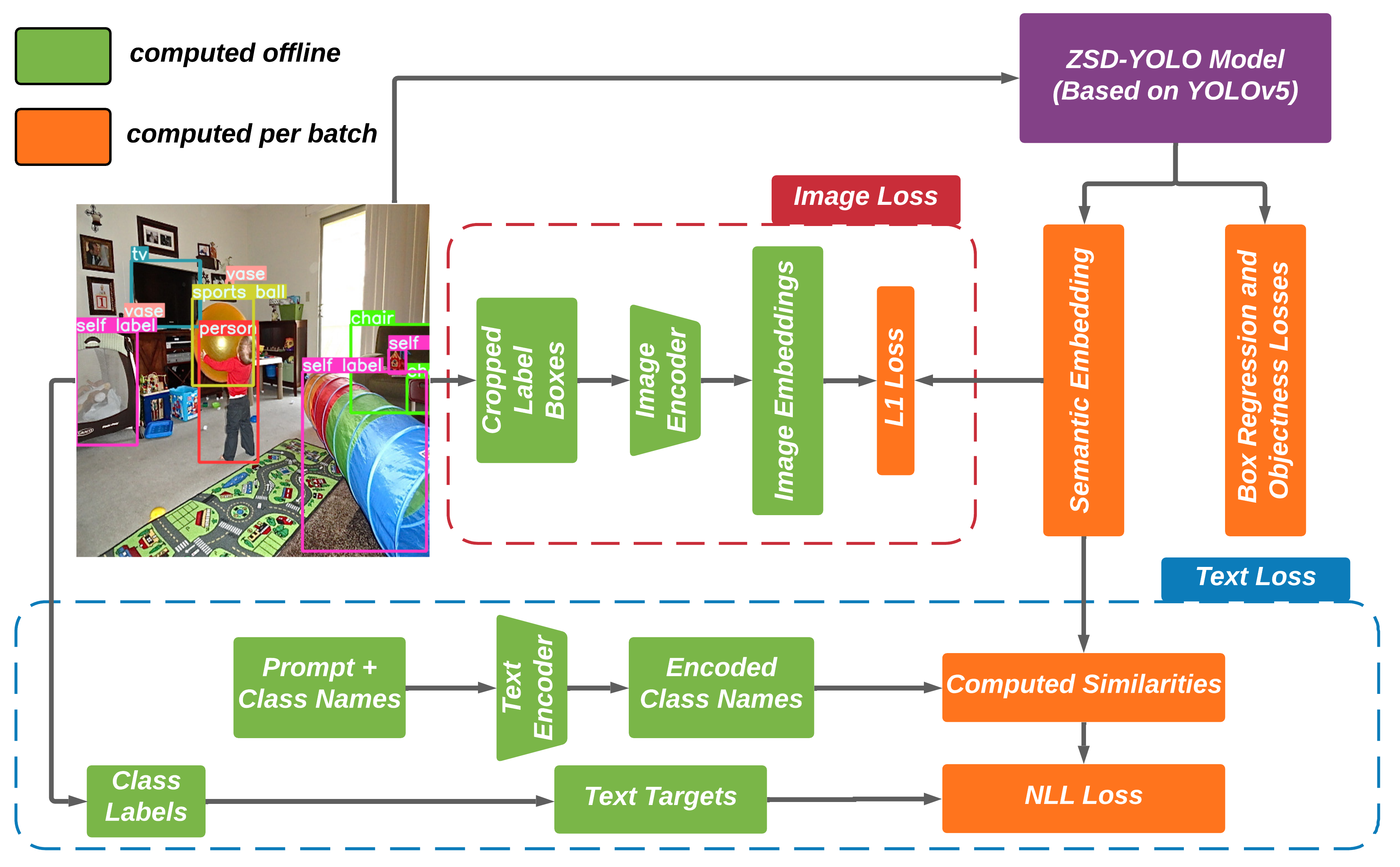}
   \caption{
   \textbf{An overview of our proposed training method of ZSD-YOLO.}
   Our method aligns detector semantic outputs to vision and language embeddings from a pretrained Vision-Language model such as CLIP.
   We modify YOLOv5 to replace typical class outputs with a semantic output with shape equal to the CLIP model embedding size and then align predicted semantic outputs of positive matched anchors with corresponding ground truth text embddings with a modified crossentropy loss described in section \ref{section:text_distill}.
   Image Embeddings of positively matched anchors are aligned using a modified $\mathcal{L}_1$ loss function described in section \ref{section:img_distill}. Best viewed in color.
   }
\label{fig:training_overview}
\vspace{-1mm}
\end{figure*}

\section{Method}
The main process for embedding alignment involves altering the YOLOv5~\cite{glenn_jocher_2022_6222936} detection head to produce a semantic embedding that mimics the CLIP model's outputs for every detection anchor.
To approximate the embedding space of CLIP with our model we minimize alignment losses for both text and image embeddings that align detector semantic outputs at each given anchor with the text and image encoder outputs of CLIP ~\cite{Radford_clip}. As shown in Figure~\ref{fig:training_overview}, we propose aligning vision and language embeddings from a model such as CLIP to adapt an one-stage detector such as YOLOv5~\cite{glenn_jocher_2022_6222936} for ZSD.

\paragraph{\textbf{Problem definition.}} The problem of ZSD as defined by \cite{Bansal_eccv_2018} involves detecting unseen classes $\mathcal{U}$ using training images $D_{train}$ containing only labeled seen classes $\mathcal{S}$ and unlabeled instances of other objects in the background denoted as $\mathcal{O}$. Note that $\mathcal{U}$, $\mathcal{S}$, $\mathcal{O}$ are pairwise disjoint. To detect novel classes, we align detector semantic outputs with the embedding space of a generalized model, in our case CLIP, by approximating its visual-semantic vector embedding space. As shown in \cite{Zheng_zsis_cvpr2021}, zero-shot prediction without some related prior knowledge is impossible, and in our case, the prior knowledge is provided by the visual-semantic embedding space of CLIP optimized for classes $\mathcal{C} = \mathcal{S}   \cup   \mathcal{U}   \cup \mathcal{O}$.\footnote{To our knowledge all previous encoders used by ZSD methods were optimized for a similarly large, general class space} During inference, our model produces semantic outputs $\mathcal{M}$ for each detection anchor that approximate the visual-semantic embedding space of CLIP by mimicking its encoders. These model semantic outputs are compared with the CLIP generated text embeddings of class names of $\mathcal{U}$ to detect instances of $\mathcal{U}$ for ZSD on the set $D_{test}$.

\vspace{-6mm}
\paragraph{\textbf{Choice of base model.}}
We choose the YOLOv5~\cite{glenn_jocher_2022_6222936} model family as our base model for three main reasons. Firstly, since typical detection models are often modified to produce only a single bounding box regression output per anchor for the task of ZSD, we believe YOLOv5~\cite{glenn_jocher_2022_6222936} is more suited for the task of ZSD as it already uses only a single bounding box regression output per anchor whereas a majority of anchor based object detectors use one bounding box regression output per class. Second, YOLOv5~\cite{glenn_jocher_2022_6222936} models and other YOLO type models generally provide the best inference speed, mAP trade-off in standard supervised object detection~\cite{Zou_detection_survey} and we find these advantages transfer well to the zero-shot detection task. Third, unlike a majority of two-stage detection models, one-stage models like YOLOv5~\cite{glenn_jocher_2022_6222936} compute class confidences for every anchor irrespective of objectness scores, a function we leverage to create a post-processing function that signficantly improves ZSD performance.

\vspace{-2mm}
\paragraph{\textbf{Pretraining procedure.}}
\label{par:pretraining}
To produce initial weights for ZSD training we train a standard YOLOv5 model of the same scale on the same seen class split with class specific annotations so that the model backbone can be initialized with class specific information when initialized for ZSD. All of these pretraining procedures are initialized with random weights and trained only on their corresponding seen label dataset. The main purpose of this process is to accelerate the training convergence rate of a zero-shot detection model.

\subsection{Aligning Text Embeddings}
\label{section:text_distill}
The process of aligning text embeddings involves the alignment of detector semantic outputs with the target text embeddings generated by CLIP~\cite{Radford_clip}. An overview of this process is highlighted in the dotted blue box found in Figure \ref{fig:training_overview}. To compute the loss for text embeddings, we first generate the seen class text embeddings $\mathcal{T}$ by feeding every seen class name into the prompt "A photo of \{class\} in the scene". The next step in computing the text loss is to extract model semantic class outputs of positive anchors $\mathcal{M}_g$ represented as "Semantic Embedding" in Figure \ref{fig:training_overview}, selected by the standard YOLOv3 \cite{Redmon/arxiv2018} procedure based on anchor box Intersection over Union (IoU) with ground-truth labels. We then generate a similarity matrix using a cosine similarity computation relating model output embeddings $\mathcal{M}_g$ and seen text embeddings $\mathcal{T}$ followed by a softmax with temperature $\tau$. We exponentiate and set $\tau$ as a learnable parameter in order to fit an optimal temperature similar to the original CLIP~\cite{Radford_clip} training, though in our final model training $\tau$ is fixed. The full computation to generate similarity matrix $\mathbf{z}$, representing the "Computed Similarities" in Figure \ref{fig:training_overview} can be written as:
\begin{align}
&\mathbf{z} = softmax\Big(\frac{\mathcal{M}_g\mathcal{T}^\top}{\lVert\mathcal{M}_g\rVert\lVert\mathcal{T}\lVert}e^\tau\Big).
\end{align}



To compute the the full text loss $\mathcal{L}_{\text{text}}$, we gather ground truth box label classes corresponding to each positively matched anchor to create one-hot encoded labels, $y$ with shape identical to the generated similarity matrix $\mathbf{z}$. The final text loss is computed with a negative log-likelihood (NLL) function measuring distance between $\mathbf{z}$ and $y$.



\subsection{Aligning Image Embeddings}
\label{section:img_distill}
The process of aligning image embeddings involves aligning model semantic outputs of each positive anchor, the same as those selected for text embedding alignment, with the corresponding image embedding. An overview of the image loss computation is outlined in the red dotted box in Figure \ref{fig:training_overview}. We first crop then preprocess each ground truth bounding box and apply the CLIP~\cite{Radford_clip} image encoder to this region offline prior to training. The preprocessing applied to the images is adapted from the official CLIP preprocessing with the typical center crop replaced with a resize. During training, all semantic outputs of anchors matched to a ground truth bounding box, the same as those selected for text-embedding alignment, are aligned with their corresponding ground-truth image embedding generated by the CLIP model.
Our final image loss function is calculated using a $\mathcal{L}_1$ loss and while we experimented with a smooth exponentiation factor for the distance function in our hyperparameter search, these did not produce any statistically significant improvements. In Figure~\ref{fig:training_overview}, let $\mathcal{I}_g$ "Image Embeddings" represent the corresponding target image embeddings. The basic image loss function is an MAE (Mean Absolute Error) function that computes distance between $\mathcal{M}_g$ and $\mathcal{I}_g$.


\subsection{Learning From Self-labeling}
\label{section:self_labeling}
To increase the number of examples our model can learn from without needing extra ground-truth labels, we apply pretrained base weights trained only on seen classes for self-labeling. To accomplish this, we run class-agnostic inference on each of the training images followed by standard post-processing and NMS on both detection outputs and ground truth labels. Ground truth labels are always prioritized over generated detections in this NMS operation to ensure that no self-label has significant overlap with a ground-truth label. During training, these self-label boxes are used to augment the data and are treated as ground truth labels for the purposes of typical one-stage detector box regression and objectness losses. We find that this allows our model to learn a more general representation of object localization by providing a wider variety of objects to learn from. For classification losses, the previous method of cropping images and applying the CLIP image encoder to the cropped patches is used along with the same loss function. Both self-label and base label examples are weighted equally in this combined loss. For text loss, these labels are not considered as there is no single corresponding class embedding that this diverse label set could be well aligned with. We visualize examples of self-labels in Figure \ref{fig:training_overview} which shows that rare object classes, not included in the COCO class label set can be identified through self-labeling allowing our model to learn from those examples. With $\mathcal{M}_s$ denoting model semantic outputs that positively match with self-label boxes and $\mathcal{I}_s$ representing the corresponding target image embeddings the full image loss function incorporating self-labels can be written as:
\begin{align}
    &\mathcal{L}_{\text{image}} = mean\Big(\Big|
    \begin{bmatrix}
    \mathcal{M}_g \\
    \mathcal{M}_s
    \end{bmatrix}
    -
    \begin{bmatrix}
    \mathcal{I}_g \\
    \mathcal{I}_s
    \end{bmatrix}
    \Big|\Big).
\end{align}

\subsection{Full Dual Loss function}
\label{section:dual_loss}
Combining both text and image alignment losses into a single function with weighting values $\mathcal{W}_t$ and $\mathcal{W}_i$ \footnote{While the total loss could be summarized with a single weighting value and an overall class weight, we find that during hyperparameter fitting, using two weight parameters is more optimal.} produces our full loss function written as:
\begin{align}
    &\mathcal{L}_{\text{dual}} = \mathcal{W}_t\mathcal{L}_{\text{text}} + \mathcal{W}_i\mathcal{L}_{\text{image}}.
\end{align}
Compared to previous ZSD works that only learn from text embedding alignment, our method is able to incorporate this dual loss and learn from both text and image embeddings. We find that this allows embeddings of the same class, which were originally all aligned to the same text embedding value, to be refined by the image loss component so that the embedding space of each class is able to gain meaningful intraclass variance. For example different models of the "car" class could be meaningfully aligned based on differences in color or type based on their image embeddings rather than all being aligned to the same text embeddding as is done in most text only approaches. The result of this process is similar to leveraging instance-wise textual descriptions~\cite{li_aaai_2019}, though our method can be applied without requiring extra data. 

\subsection{Inference on Unseen classes}
During inference, the similarity matrix $\mathbf{z}$ is computed in the same manner as during training, but with reference embeddings $\mathcal{T}$ set equal to the unseen class name embeddings. We also explore modifying typical YOLO post-processing in order to reduce typical bias of seen objects being given high confidence scores during zero-shot detection and generalized zero-shot detection as they tend to have higher objectness scores.

\paragraph{\textbf{Text embeddings during inference.}}
When analyzing the cosine similarity between image and text embeddings of certain unseen classes, we found that CLIP often embeds a word incorrectly when only an ambiguous class name is used. \footnote{This limitation was also found by the original CLIP authors who mentioned this in their code repository: https://github.com/openai/CLIP} For example, the unseen class "mouse" embedding was more semantically similar to the animal mouse, and not the computer mouse, the correct version. Prior ZSD methods avoided this issue by selecting the correct word embedding, based on the WordNet~\cite{wordnet} synset, from word vectorizers such as word2vec~\cite{word2vec}, but the CLIP text embedding does not support this approach. Therefore, we sought a simple, automatic method to correct unseen text reference embeddings that could be applied to any prompt class. To this end, during inference we also include the WordNet~\cite{wordnet} definition of the class with the prompt "a photo of \{class\}, \{definition\}, in the scene." The inclusion of each class definition not only produces more semantically accurate unseen class reference embeddings, but also distanced previously similar classes by providing specificity. For example, cosine distance between unseen classes "bear" and "cat" increased significantly with the addition of definitions allowing for their class APs to improve by reducing incorrect detections between the two similar classes. While this method does provide an overall score increase, not all class APs improve. Still, we benchmark only applying these evenly across all unseen classes, as benchmarking must adhere to these constraints. In the supplemental materials, we further discuss the results of this process.

\paragraph{\textbf{Post-processing.}}
\label{par:post-processing}
To generate predictions, we first apply a very low objectness cutoff which rarely eliminates unseen class examples compared to two stage approaches. For all remaining anchors, we create a confidence value by multiplying their objectness scores by the maximum class confidence value of each prediction. We then apply a cutoff value to these generated confidence scores. After applying NMS, the final confidence score given to these remaining predictions is only the maximum calculated class similarity value and does not factor in objectness which typically favors seen objects. For our final predictions, we apply NMS and limit max detections because our method tends to produce many high confidence predictions. 
\section{Experiments}
We compare our proposed method with previous methods under the standard zero-shot detection (ZSD) and generalized zero-shot detection (GZSD) settings. In addition, we present ablation studies to isolate the improvements from each component in our proposed method

\subsection{Dataset}
\label{section:dataset}
We mainly evaluate model performance under the two most common ZSD benchmarks which use the COCO~\cite{Lin_TY_COCO_2014} dataset under the 65/15~\cite{Rahman_aaai_2020} and 48/17~\cite{Bansal_eccv_2018} ZSD class splits. Both training sets are generated by removing images from the COCO 2014 training set that contain any unseen object instances and sampling the validation images that contain at least one unseen image. To test our model's ability to generalize across datasets we also benchmark on the established ZSD class split for ILSVRC~\cite{ILSVRC15} proposed in Rahman~\etal~\cite{Rahman_accv_2018} and the Visual Genome~\cite{krishna_visualgenome} class split proposed in Bansal~\etal~\cite{Bansal_eccv_2018}. Since the previous state-of-the-art~\cite{Zheng_zsis_cvpr2021} did not benchmark on these two datasets we compare with the best performing previous method for these benchmarks. To avoid optimizing training hyperparameters and methods on any of the standard ZSD testing sets, we create a ZSD validation set. To create this validation set, we take all training images with unseen annotations from the 65/15 COCO split, the unused part of the original COCO 2014 training set, resulting in a validation set with 20,483 images. 

\subsection{Evaluation Protocol}
We benchmark the ZSD and GZSD performance on COCO 48/17 split and 65/15 split. In ZSD, only unseen text embeddings are present during inference and the model predicts only unseen object instances. We report mAP@0.5 and Recall@100 at various IOU thresholds. In GZSD, both seen and unseen text embeddings are present in validation and the model predicts both seen and unseen object instances. Additionally, for GZSD we also report the harmonic mean (HM) of seen and unseen mAP@0.5 and Recall@100 at three IOU thresholds. Similar to previous ZSD papers~\cite{Zheng_zsis_cvpr2021}, mAP will always refer to mAP@0.5 unless otherwise specified. For the ILSVRC benchmark, we measure ZSD performance using mAP on the 23 unseen classes. For Visual genome, ZSD performance is measured by averaging the Recall@100 scores of 130 unseen classes at three given IOU scores because the Visual Genome dataset is known to have many missing labels due to the large class set. All benchmarking outside of ablation studies use the proposed self-labeling (\ref{section:self_labeling}), dual loss function (\ref{section:dual_loss}), and ZSD tailored post-processing (\ref{par:post-processing}).

\subsection{Implementation Details}
Our code implementation is based upon the Pytorch YOLOv5 repository by ultralytics.\footnote{https://github.com/ultralytics/yolov5} For model architecture, modifications are made to the final convolutional prediction layer to change class outputs to match the 512 CLIP embedding outputs from the publicly available ViT-B/32 model.\footnote{https://github.com/openai/CLIP} For loss we only alter class side losses. To fit hyperparameters, we run hyperparameter search with the COCO 65/15 split training set and use the unseen mAP of our proposed validation set from~\ref{section:dataset} as our fitness objective for approximately 200 evolution generations. Our main loss hyperparameters are a text loss weight of 1.05, image loss weight of 1.21, box loss weight of 0.03, class loss weight of 0.469, and objectness loss weight of 2.69. Our main SGD optimizer hyperparameters are base lr of 0.00282, momentum of 0.854, weight decay of 0.00038. Our models are trained on a 50 epoch cosine schedule and initialized with pretrained weights described in Section~\ref{par:pretraining}.

\begin{table}[!t]
\caption{ZSD results on two COCO splits. Seen/Unseen refers to the dataset splits.}
\label{table:ZSD_benchmark}
\begin{center}
\resizebox{0.9\linewidth}{!}{
\begin{tabular}{c|c|c|c|c|c}
\toprule
\multirow{2}{*}{Method} & \multirow{2}{*}{Seen/Unseen} & \multicolumn{3}{c}{Recall@100} & mAP \\
\cmidrule(lr){3-5} \cmidrule(lr){6-6}
~ & ~ & 0.4 & 0.5 & 0.6 & 0.5 \\
\midrule
SB~\cite{Bansal_eccv_2018} & 48/17 & 34.5 & 22.2 & 11.4 & 0.4 \\
DSES~\cite{Bansal_eccv_2018} & 48/17 & 40.3 & 27.2 & 13.7 & 0.6 \\
TD~\cite{li_aaai_2019} & 48/17 & 45.6 & 34.4 & 18.2 & - \\
PL~\cite{Rahman_aaai_2020} & 48/17 & - & 43.6 & - & 10.1 \\ 
ZSD-CNN-ohem~\cite{wang_ictai_2020} & 48/17 & 47.8 & 41.2 & 34.4 & -\\
Gtnet~\cite{zhao_aaai_2020} & 48/17 & 47.4 & 44.7 & 35.6 & - \\ 
DELO~\cite{zhu2020don} & 48/17 & - & 33.6 & - & 7.6 \\ 
BLC~\cite{zheng_accv_2020} & 48/17 & 49.7 & 46.4 & 41.9 & 9.9\\
ZSI~\cite{Zheng_zsis_cvpr2021} & 48/17 & 57.4 & 53.9 & \textbf{48.3} & 11.4 \\
\hline
ZSD-YOLOx & 48/17 & \textbf{61.6} & \textbf{55.8} & 46.2 & \textbf{13.4} \\
\bottomrule
PL~\cite{Rahman_aaai_2020} & 65/15 & - & 37.8 & - & 12.4  \\
BLC~\cite{zheng_accv_2020} & 65/15 & 54.2 & 51.7 & 47.9 & 13.1 \\
ZSI~\cite{Zheng_zsis_cvpr2021} & 65/15 & 61.9 & 58.9 & 54.4 & 13.6 \\
\hline
ZSD-YOLOx & 65/15 & \textbf{75.5} & \textbf{69.5} & \textbf{57.3} & \textbf{18.3} \\
\bottomrule
\end{tabular}}
\end{center}
\vspace{-4mm}
\end{table}
\begin{table}[tbp]
\caption{Comparison of our method with the previous GZSD methods on two COCO splits. HM denotes the harmonic mean for seen and unseen class split mAP and Recall@100.}
\label{table:GZSD_benchmark}
\begin{center}
\resizebox{\linewidth}{!}{
\begin{tabular}{cccccccc}
\toprule
\multirow{2}{*}{Method} & \multirow{2}{*}{Seen/Unseen} & \multicolumn{2}{c}{Seen} & \multicolumn{2}{c}{Unseen} & \multicolumn{2}{c}{HM} \\
\cmidrule(lr){3-4} \cmidrule(lr){5-6} \cmidrule(lr){7-8}
~ & ~ & mAP & Recall@100 & mAP & Recall@100 & mAP & Recall@100 \\
\midrule
DSES~\cite{Bansal_eccv_2018} & 48/17 & - & 15.1 & - & 15.4 & - & 15.2 \\
PL~\cite{Rahman_aaai_2020} & 48/17 & 36.0 & 38.3 & 4.2 & 26.4 & 7.4 & 31.2 \\
BLC~\cite{zheng_accv_2020} & 48/17 & 42.2 & 57.6 & 4.6 & 46.4 & 8.3 & 51.4 \\
ZSI~\cite{Zheng_zsis_cvpr2021} & 48/17 & \textbf{46.6} & \textbf{70.8} & 4.9 & \textbf{53.9} & 8.8 & \textbf{61.2} \\
\hline
ZSD-YOLOv5x & 48/17 & 31.7 & 63.3 & \textbf{13.6} & 45.2 & \textbf{19.0} & 52.7 \\
\bottomrule
PL~\cite{Rahman_aaai_2020} & 65/15 & 34.1 & 36.4 & 12.5 & 37.2 & 18.2 & 36.8 \\
BLC~\cite{zheng_accv_2020} & 65/15 & 36.1 & 56.4 & 13.2 & 51.65 & 19.3 & 54.0  \\
ZSI~\cite{Zheng_zsis_cvpr2021} & 65/15 & \textbf{38.7} & \textbf{67.2} & 13.7 & 59.0 & 20.2 & 62.8  \\
\hline
ZSD-YOLOv5x & 65/15 & 31.7 & 61.0 & \textbf{17.9} & \textbf{65.2} & \textbf{22.9} & \textbf{63.0}\\
\bottomrule
\end{tabular}}
\end{center}
\vspace{-5mm}
\end{table}

\begin{table*}[t]
  \caption{ZSD class AP for unseen classes of ILSVRC DET 2017 dataset.} 

  \begin{center}\setlength\tabcolsep{2pt}
  \scalebox{0.8}{
    \begin{tabular}{c|c||c|c|c|c|c|c|c|c|c|c|c|c|c|c|c|c|c|c|c|c|c|c|c}
    \toprule[0.1em]
    
\rotatebox{90}{ }&\rotatebox{90}{ \textbf{mean}}&\rotatebox{90}{p.box}&\rotatebox{90}{syringe}&\rotatebox{90}{harmonica}&\rotatebox{90}{maraca}&\rotatebox{90}{burrito}&\rotatebox{90}{pineapple}&\rotatebox{90}{electric-fan}&\rotatebox{90}{iPod}&\rotatebox{90}{dishwasher}&\rotatebox{90}{canopener}&\rotatebox{90}{plate-rack}&\rotatebox{90}{bench}&\rotatebox{90}{bowtie}&\rotatebox{90}{s.trunk}&\rotatebox{90}{scorpion}&\rotatebox{90}{snail}&\rotatebox{90}{hamster}&\rotatebox{90}{tiger}&\rotatebox{90}{ray}&\rotatebox{90}{train}&\rotatebox{90}{unicycle}&\rotatebox{90}{golfball}&\rotatebox{90}{h.bar}\\  \toprule[0.1em]

SAN\cite{Rahman_accv_2018} &16.4 &5.6 &1.0 &0.1 &0.0 &27.8 &1.7& 1.5 &1.6 &7.2& 2.2& 0.0& 4.1& 5.3& \textbf{26.7} & 65.6& 4.0 &47.3& 71.5& 21.5 &51.1& 3.7 &26.2& 1.2 \\
 DSES\cite{Bansal_eccv_2018} &22.7&7.4&2.3&1.1&0.6&46.2&4.3&8.7&32.7&14.6&6.9&9.1&7.4&4.9&6.9&73.4&7.8&56.8&80.8&24.5&59.9&25.4&33.1&7.6\\
 ZSDTD\cite{li_aaai_2019} &24.1&7.8&3.1&1.9&1.1&49.4&4.0&9.4&35.2&14.2&8.1&10.6&9.0&\textbf{5.5}&8.1&\textbf{73.5}&8.6&57.9&\textbf{82.3}&\textbf{26.9}&61.5&24.9&38.2&8.9 \\ 
 GTNet\cite{zhao_aaai_2020} &26.0&4.4&\textbf{30.4}&2.3&1.2&\textbf{51.5}&5.2&18.5&\textbf{40.6}&18.0&13.1&4.7&\textbf{13.7}&4.6&19.2&69.7&10.2&74.7&72.7&1.4&\textbf{65.7}&27.1&\textbf{40.4}&\textbf{9.1} \\ 
\hline
ZSD-YOLOv5x & \textbf{27.8} & \textbf{24.0} & 24.4 & \textbf{6.4} & \textbf{2.2} & 45.1 & \textbf{28.1} & \textbf{32.0} & 24.5 & \textbf{26.2} & \textbf{26.0} & \textbf{13.1} & 11.7 & 4.9 & 17.3 & 72.9 & \textbf{33.3} & \textbf{78.1} & 57.9 & 0.3 & 37.3 & \textbf{37.6} & 32.8 & 4.0 \\ 

\bottomrule[0.1em]

    \end{tabular}}
  \end{center}
  \label{tab:ilsvrc}
\vspace{-5mm}
\end{table*}

\begin{table}[ht]
\caption{Recall@100 comparison of our method with the previous state-of-the-art ZSD works on Visual Genome~\cite{krishna_visualgenome}.}
\label{table:VG_benchmark}
\begin{center}
\resizebox{0.7\linewidth}{!}{
\begin{tabular}{c|c|c|c}
\toprule
\multirow{2}{*}{Method} & \multicolumn{3}{c}{Recall@100} \\
\cmidrule(lr){2-4}
~ & 0.4 & 0.5 & 0.6 \\
\midrule
SAN~\cite{Rahman_accv_2018} & 6.8 & 5.9 & 3.1 \\
LAB~\cite{Bansal_eccv_2018} & 8.4 & 5.4 & 2.7 \\
ZSDTD~\cite{li_aaai_2019} & 9.7 & 7.2 & 4.2 \\
ZSD-CNN-ohem~\cite{wang_ictai_2020} & 13.7 & 11.0 & 8.3 \\
GTNet\cite{zhao_aaai_2020} & 14.3 & 11.3 & 8.9 \\
\hline
ZSD-YOLOv5x & \textbf{27.8} & \textbf{23.0} & \textbf{15.1} \\
\bottomrule
\end{tabular}}
\end{center}
\vspace{-4mm}
\end{table}
\subsection{Comparison with State-of-the-art}
\paragraph{COCO}
We compare our results under the ZSD task for the two COCO ZSD splits in Table~\ref{table:ZSD_benchmark}. We observe that under the 48/17 split, our method improves over the previous state-of-the-art approach~\cite{Zheng_zsis_cvpr2021} by 2.0 mAP, a 17.6\% improvement. Under the 65/15 split our method surpasses the best scoring previous ZSD method~\cite{Zheng_zsis_cvpr2021} by 4.7 mAP or approximately a 34.6\% improvement. Generally, our method performs better under the 65/15 split mainly due because our model is able to approximate the CLIP embedding space better with a more diverse training label set.

Our results for the GZSD COCO task are listed in Table~\ref{table:GZSD_benchmark}. Under the GZSD setting, we focus our method development on Unseen mAP and HM (Harmonic Mean) mAP and our method surpasses the previous state-of-the-art~\cite{Zheng_zsis_cvpr2021} in both metrics and under both splits. In the 48/17 split our method improves by 8.7\% mAP, or 177.6\% under unseen mAP and 10.2 mAP or 116.0\% for HM mAP, the main GZSD benchmark. In the 65/15 split our method improves by 4.2 unseen mAP, or a 30.7\% improvement. For Harmonic Mean (HM) mAP, our method improves by 2.7 mAP or a 13.4\% improvement. Through ablation experiments (\ref{par:post-processing_ablations}), we find that a large factor in our strong GZSD benchmark scores is due to our proposed post-processing function (\ref{par:post-processing}).

\paragraph{ILSVRC}
Under the ILSVRC split, shown in Table~\ref{tab:ilsvrc}, many other methods such as DSES\cite{Bansal_eccv_2018}, ZSDTD~\cite{wang_ictai_2020}, and GtNet\cite{zhao_aaai_2020} use additional data or information not present in typical ZSD benchmarking. Despite these advantages that our method does not leverage, we still surpass the previous state-of-the-art by 1.8 mAP or approximately a 6.9\% improvement.

\paragraph{Visual Genome}
Unlike the COCO and ILSVRC dataset splits, Visual Genome, shown in Table~\ref{table:VG_benchmark} is benchmarked only using Recall@100 for three given IOU values because the large class number leads to many missing labels. Our method significantly improves over previous ones. On Recall@100 for IOU=0.5 our method improves over the previous state-of-the-art by 11.7,  more than doubling the result~\cite{zhao_aaai_2020}. This benchmark shows that our method can improve over other methods far more significantly on larger datasets with a greater number of seen classes to learn from. Additionally, the Visual Genome benchmark shows the strength of our post-processing approach which tends to recall unseen objects far better than typical post-processing as it does not eliminate as many unseen objects through objectness score cutoffs.

\subsection{Inference Speed Comparison}
\label{section:speed_benchmarking}

\paragraph{\textbf{Benchmark settings.}} Model speeds are evaluated on both single image throughput and mini-batch throughput with batch size determined by the GPU memory utilization of each model. We measure all latencies on the 65/15 dataset split with a single V100 GPU as the listed ZSD approaches have negligible inference time differences between splits. The ZSD-YOLOv5l model is missing from our scaling experiments for reasons explained in our self-labeling ablation study. We also implement our method using the YOLOv3 base detector, and list it as ZSD-YOLOv3.
In evaluating inference latency across all models, we include only the amount of time it takes for the model to produce outputs on the batch which includes the calculations of cosine similarities and the softmax with temperature operation, but does not include post-processing operations.


\paragraph{\textbf{Results.}} We display the results of our model scaling and compare it to the previous state-of-the-art approach~\cite{Zheng_zsis_cvpr2021}, in Table~\ref{table:scaling_benchmark}. Compared to the previous state-of-the-art~\cite{Zheng_zsis_cvpr2021}, our best model, ZSD-YOLOv5x, is over 3x faster on single image inference while also being 4.7 mAP better. For a closer model strength comparison, we train a YOLOv3 model with our method as the strength of the base YOLOv3 model is most similar to the base detector used in ZSI~\cite{Zheng_zsis_cvpr2021} and other ZSD methods and find that our method is still able to outperform ZSI by 3.2 mAP on the 65/15 dataset split while running approximately 9.7x faster on single image inference. To measure scaling performance, we focus on the 48/17 split as our experimentation in developing our method under the 65/15 split likely resulted in certain model scales being slightly more favored in the specific split. In supervised YOLOv5 detection, YOLOv5m and YOLOv5x improve over YOLOv5s by 13.9\% and 24.2\% respectively in terms of mAP~\cite{glenn_jocher_2022_6222936} whereas our proposed ZSD-YOLOv5m and ZSD-YOLOv5x improve over ZSD-YOLOv5s by 14.0\% and 34.0\% in terms of mAP. Given that our relative mAP improvement percentages from scaling are similar to that of the original YOLOv5 models, we conclude that our method scales similarly to fully supervised YOLOv5 detectors. 

\begin{table}[!t]
\caption{Comparison of our method's inference latency with previous the state-of-the-art~\cite{Zheng_zsis_cvpr2021} on 65/15 and 48/17 COCO splits. Batch sizes for mini-batch throughput are listed in parentheses in column 3.}
\label{table:scaling_benchmark}
\begin{center}
\resizebox{\linewidth}{!}{
\begin{tabular}{cccccccc}
\toprule
\multirow{2}{*}{Method} & \multicolumn{2}{c}{Latency (ms)} & \multicolumn{2}{c}{48/17} & \multicolumn{2}{c}{65/15} \\
\cmidrule(lr){2-3} \cmidrule(lr){4-5} \cmidrule(lr){6-7}
~ & single & max (batch size) & mAP & Recall & mAP & Recall \\
\midrule
ZSI~\cite{Zheng_zsis_cvpr2021} & 162.0 & 100.1 (4) & 11.6 & 54.9 & 13.6 & 58.9 \\
\hline
ZSD-YOLOv3 & 16.7 & 4.6 (20) & 11.2 & 54.1 & 16.8 & 66.1 \\
ZSD-YOLOv5s & 15.5 & 1.7 (64) & 10.0 & 47.9 & 14.0 & 60.4 \\
ZSD-YOLOv5m & 19.1 & 2.7 (32) & 11.4 & 53.2 & 17.4 & 65.9 \\
ZSD-YOLOv5x & 51.4 & 6.4 (16) & 13.4 & 55.8 & 18.3 & 69.5 \\
\bottomrule
\end{tabular}}
\end{center}
\vspace{-10mm}
\end{table}



\begin{table*}[t]
\begin{center}
\caption{COCO ZSD split results of our ablation study on our proposed self-labeling and post-processing methods. Note that the baseline method in this study uses both image and text losses as we ablate the image loss component separately in Table~\ref{table:image_ablation}. All results are produced using the ZSD-YOLOv5x sized model.}
\label{table:ablation}
\resizebox{1.0\linewidth}{!}{
\begin{tabular}{c|ccccccccccccc}
\toprule
\multirow{4}{*}{ } & \multirow{4}{*}{ZSD} & \multirow{4}{*}{self-labeling} & \multirow{4}{*}{ZSD-POST} & \multicolumn{4}{c}{ZSD} & \multicolumn{6}{c}{GZSD} \\
\cmidrule(lr){5-8} \cmidrule(lr){9-14}
~ & ~ & ~ & ~ & \multicolumn{4}{c}{Unseen} & \multicolumn{2}{c}{Seen} & \multicolumn{2}{c}{Unseen} & \multicolumn{2}{c}{HM} \\
\cmidrule(lr){5-8} \cmidrule(lr){9-10} \cmidrule(lr){11-12} \cmidrule(lr){13-14}
~ & ~ & ~ & ~ & mAP & \multicolumn{3}{c}{Recall@100} & mAP & Recall@100 & mAP & Recall@100 & mAP & Recall@100 \\
\cmidrule(lr){5-5} \cmidrule(lr){6-8}  \cmidrule(lr){9-9} \cmidrule(lr){10-10} \cmidrule(lr){11-11} \cmidrule(lr){12-12} \cmidrule(lr){13-13} \cmidrule(lr){14-14}
~ & ~ & ~ & ~ & 0.5 & 0.4 & 0.5 & 0.6 & 0.5 & 0.5 & 0.5 & 0.5 & 0.5 & 0.5\\
\hline
\multirow{4}{*}{\rotatebox{90}{48/17}} & \checkmark & & & 10.2 & 57.5 & 53.2 & 47.4 & 29.7 & 57.9 & 9.3 & 43.9 & 14.2 & 50.3 \\ 
~ & \checkmark & \checkmark & & 11.8 & 60.1 & 55.6 & \textbf{49.8} & 32.4 & 63.2 & 12.7 & 44.3 & 18.2 & 52.1\\
~ & \checkmark & & \checkmark & 12.6 & 60.2 & 54.9 & 45.0 & \textbf{34.6} & \textbf{65.6} & 11.1 & \textbf{51.4} & 16.8 & \textbf{57.6}\\
~ & \checkmark & \checkmark & \checkmark & \textbf{13.4} & \textbf{61.6} & \textbf{55.8} & 46.2 & 31.7 & 63.3 & \textbf{13.6} & 45.2 & \textbf{19.0} & 52.7 \\
\hline 
\multirow{4}{*}{\rotatebox{90}{65/15}} & \checkmark & & & 13.7 & 69.1 & 64.8 & 58.6 & 30.6 & 57.2 & 13.7 & 62.1 & 18.9 & 59.6 \\
~ & \checkmark & \checkmark & & 14.7 & 73.8 & 69.4 & \textbf{62.4} &  \textbf{36.3} & \textbf{66.0} & 14.7 & \textbf{68.3} & 20.9 & \textbf{67.2}\\
~ & \checkmark & & \checkmark & 16.6 & 73.9 & 67.9 & 56.7 & 32.7 & 61.1 & 16.6 & 66.4 & 22.0 & 63.6\\
~ & \checkmark & \checkmark & \checkmark & \textbf{18.3} & \textbf{75.5} & \textbf{69.5} & 57.3 & 31.7 & 61.0 & \textbf{17.9} & 65.2 & \textbf{22.9} & 63.0\\
\hline
\end{tabular}
}
\end{center}
\vspace{-8mm}
\end{table*}

\subsection{Ablation Study}
We present the effects of our proposed self-labeling and post-processing method in Table~\ref{table:ablation}, and study the effect of our image loss component seperately in Table~\ref{table:image_ablation}. In developing our method, we follow previous ZSD literature which uses the unseen mAPs for both ZSD and GZSD as the primary benchmarks and we find using both self-labeling and our proposed post-processing yields the best results across benchmark splits and tasks.
\paragraph{Self-labeling}

Comparing the third and fourth rows of each COCO ZSD split in Table~\ref{table:ablation}, we find that self-labeling provides an overall improvement of 0.8 mAP and 1.7 mAP for the 48/17 and 65/15 class splits respectively. Through experimentation, we find that a model which labels using its own base weights tends to create self-labels it has already learned through pretraining and therefore does not "expand" its knowledge of what can be considered an object. For example, we find that when the ZSD-YOLOv5x sized model is self-labeled with ZSD-YOLOv5l, the mAP gain is 0.6 mAP and 0.8 mAP greater, for the 48/17 and 65/15 class splits respectively, than when self-labeled with its own base weights. Therefore, we self-label using the ZSD-YOLOv5l sized detector in our ablation studies and scaling experiments where we do not test the ZSD-YOLOv5l model (\ref{section:speed_benchmarking}).

\paragraph{Post-processing}
\label{par:post-processing_ablations}
When studying the effect of our post-processing function, we consider row two of each ZSD split in Table \ref{table:ablation} to be the baseline which only uses typical YOLOv5 ZSD post-processing, whereas row four uses our proposed post-processing. We find that our proposed post-processing provides a mAP improvement of 1.6 mAP and 3.6 mAP on the 48/17 and 65/15 COCO ZSD splits respectively. Though when benchmarking GZSD, we find that while our method provides a similar significant unseen class improvement, we also find a loss in seen class detection mAP which is expected as our post-processing method favors unseen classes by reducing the influence of objectness scores when computing detection confidence. Therefore, since our proposed processing has a negligible impact on inference latency, we use our proposed post-processing throughout all of our other experiments. The strength of this post-processing is also another reason we opted for a one stage detection approach as a two stage detection approach cannot consider the class confidences of every anchor due to computational constraints and must first filter out anchors proposals with low objectness scores before computing class scores which we find leads to filtering out unseen object instances.

\vspace{-2mm}
\paragraph{Image loss component}
We study the effect of the image loss component of our loss function by training with image loss suppressed and other hyperparameters the same as the dual loss training. Our results are compiled in Table~\ref{table:image_ablation}. We find that combining the image loss with the text loss using our proposed dual loss function (ZSD-YOLOv5x-dual) improves ZSD performance by approximately 4.4 mAP on the 48/17 split and 1.5 mAP on the 65/15 split over the text-only baseline (ZSD-YOLOv5x-text). Considering the improvement on the 48/17 split is far greater than the improvement on the 65/15 split, we conclude that the effect additional image embedding loss is accentuated in limited data scenarios but is still able to provide a significant improvement even when there is sufficient data.

\begin{table}[!t]
\caption{Comparison of image loss component of our described dual loss function when used to train ZSD-YOLOv5x on the 65/15 and 48/17 COCO dataset splits. We find that using both image and text losses (dual loss) provides optimal results.}
\vspace{-4mm}
\label{table:image_ablation}
\begin{center}
\resizebox{\linewidth}{!}{
\begin{tabular}{c|c|c|c|c|c}
\toprule
\multirow{2}{*}{Method} & \multirow{2}{*}{Seen/Unseen} & \multicolumn{3}{c}{Recall@100} & mAP \\
\cmidrule(lr){3-5} \cmidrule(lr){6-6}
~ & ~ & 0.4 & 0.5 & 0.6 & 0.5 \\
\midrule
ZSD-YOLOv5x-text & 48/17 & 58.8 & 53.4 & 45.0 & 9.0\\
ZSD-YOLOv5x-dual & 48/17 & \textbf{61.6} & \textbf{55.8} & \textbf{46.2} & \textbf{13.4} \\
\bottomrule
ZSD-YOLOv5x-text & 65/15 & 71.4 & 66.0 & 56.4 & 16.8 \\
ZSD-YOLOv5x-dual & 65/15 & \textbf{75.5} & \textbf{69.5} & \textbf{57.3} & \textbf{18.3} \\
\bottomrule
\end{tabular}}
\end{center}
\vspace{-6mm}
\end{table}



\vspace{-2mm}

\section{Conclusion}
We introduce ZSD-YOLO, a zero-shot detector that surpasses all previous ZSD results on the main ZSD benchmarks using the COCO\cite{Lin_TY_COCO_2014}, ILSVRC\cite{ILSVRC15}, and Visual Genome~\cite{krishna_visualgenome} datasets. Furthermore, we devise a self-labeling method that can improve ZSD performance without needing new data or labels and optimize single stage post-processing for the ZSD task. We explore the effect of traditional scaling approaches on the ZSD task and find that typical model scaling can transfer well to our method to create a family of efficient and accurate ZSD models. ZSD is an emerging research direction and has many potential applications in computer vision such as a generalized object detector. We hope that our work can establish a strong ZSD baseline that utilizes image and text loss components to inspire further work in the crucial ZSD domain.



{
    \clearpage
    \bibliographystyle{ieee_fullname}
    \bibliography{macros,main}
}
\clearpage


\appendix

\section{Supplementary Introduction}

In the supplementary materials we provide the following items:
\begin{enumerate}
\item Clarification on the differences between zero-shot detection, open-vocabulary detection, and weakly supervised detection and why our method is considered zero-shot detection in Section \ref{Zero-shot Detection vs Open-vocabulary Detection vs Weakly Supervised Detection}.
\item Results of the definition ablation study describing the exact effects of adding definitions to inference text embedding prompts in Section \ref{Definition Ablation}.
\item Details on our proposed validation set upon which we develop our methods and fit hyperparameters in Section \ref{Validation Set Details}
\item Discussion of qualitative results demonstrating both the strong performance of our model as well as examples of common mistakes in Section \ref{Qualitative Results}.
\item More information on implementation details including our hyperparameter search process, post-processing constants, and self-labeling constants in Section \ref{Further Implementation Details}.
\item An additional ablation study on the Visual Genome~\cite{krishna_visualgenome} ZSD split\cite{Bansal_eccv_2018} justifying our choice of post-processing in Section \ref{Visual Genome Ablation}.

\end{enumerate}
\section{Zero-shot Detection vs Open-vocabulary Detection vs Weakly Supervised Detection}
\label{Zero-shot Detection vs Open-vocabulary Detection vs Weakly Supervised Detection}
Given the similarities between the tasks of zero-shot detection (ZSD),  open-vocabulary detection (OVD), and weakly supervised detection (WSD), we would like to clarify the differences and explain why we have classified our method as ZSD. While these tasks all address the issue of learning to detect objects using limited data, each provide different constraints on the type of training data that can be used.  

Firstly, weakly supervised detection, arguably the outlier in this suite of tasks, uses only image-labels to train a detector. Therefore, the main challenge in WSD~\cite{bilen2016weakly,wan2019c,cinbis2016weakly} is localizing novel objects since novel classes are often present in the training data. WSD approaches generally also require knowledge of the target novel classes during training, whereas ZSD and OVD approaches must be able to target any subset of the entire language vocabulary.

In comparing OVD and ZSD, ZSD is the more restrictive task, as acknowledged by Zareian~\etal~\cite{Zareian/cvpr2021}. The main difference between the OVD benchmarks and ZSD benchmarks are that ZSD training datasets eliminate all training images with instances of unseen objects whereas OVD benchmarks do not.\footnote{With the exception of the Visual Genome ZSD benchmark for reasons explained in~\cite{Bansal_eccv_2018}} While the labels of unseen objects are removed, OVD models are still able to learn from these examples. Most recently the method proposed by Gu~\etal~\cite{gu2021openvocabulary} distills vision-language information to detect unseen classes. The method of Gu~\etal~\cite{gu2021openvocabulary} is able to learn from background examples through their proposed image distillation method. Using only text distillation, they achieve only 5.9 mAP\footnote{Similar to our main paper, mAP refers to mAP@0.5} \cite{gu2021openvocabulary} on unseen classes whereas using only image distillation they achieve 24.1 mAP~\cite{gu2021openvocabulary} on the 48/17 COCO OVD benchmark from~\cite{Zareian/cvpr2021}. In ablation studies, they demonstrated that the contribution of the image and text losses were approximately the same under their proposed LVIS~\cite{gupta2019lvis} benchmark. Note that our text only method is still able to achieve 9.0 mAP under the 48/17 ZSD benchmarking split which eliminates all training images with unseen samples. Given these results, we conclude that the inclusion of unseen objects in the background of images provides an enormous advantage to OVD methods. The authors of ~\cite{gu2021openvocabulary} also acknowledge this fact in discussing their COCO 48/17 OVD~\cite{Zareian/cvpr2021} results. With this in mind, ZSD methods must be able to perform detection on classes that are not present in training images at all whereas OVD methods require the inclusion of unseen objects within their training sets to perform well. Therefore, when unseen objects are unavailable, as is often the case since one could not simply just include unlabeled images in their object detection training sets, OVD models fail to perform well in detecting those entirely unseen object types. 

While Zareian~\etal~\cite{Zareian/cvpr2021} states the usage of image captioning data classifies a method as OVD, previously published works~\cite{li_aaai_2019,zhang2020zero} have used similar image captioning while still being classified as ZSD. For this reason, we believe the usage of CLIP~\cite{Radford_clip}, which is trained on a large corpus of image text pairs can still be considered a ZSD method. Additionally, CLIP has been considered a zero-shot model since its introduction and in the various downstream tasks leveraging its embedding space. Furthermore, the CLIP embedding space is optimized for a vocabulary of similar size compared to other embedding spaces used such as GloVe~\cite{pennington-etal-2014-glove} and word2vec~\cite{word2vec}. Therefore, in terms of application, our method can transfer to a similar number of novel class names. The CLIP encoders also support full sentence prompts as demonstrated by our inference prompts including definitions allowing for nearly any prompt to be used for detection, thus our method can be applied just as widely, if not more widely than previous ZSD methods. While we concede that the CLIP embedding space may have stronger visual-semantic mappings, we cannot apply our method using previous word vectorizers to directly compare with their results. Similarly, we are unsure how to apply previous methods using CLIP embeddings and therefore cannot in good faith attempt to implement their methods to draw a comparison.

\section{Definition Ablation}
\label{Definition Ablation}
To study the direct effects of adding definitions to inference prompts on each class, we present the class-wise APs with and without definitions in inference prompts for the COCO 48/17~\cite{Bansal_eccv_2018} and 65/15~\cite{Rahman_aaai_2020} ZSD splits in Table \ref{tab:COCO_48_17_def_ablation} and Table \ref{tab:COCO_65_15_def_ablation} respectively. Following the analysis in our main paper, the "bear" and "cat" class APs had the largest increase under the 65/15 benchmarking split, increasing by 175.1\% and 120.0\% respectively, given that their embedding cosine similarities decreased from 0.924 to 0.775 with the addition of definitions. The case of these two classes demonstrates that the inclusion of definitions can aid in differentiating between two previously similar text embeddings and reduce the number of false positives between the two classes. Furthermore, other classes such as the "mouse" class that were previously mapped incorrectly due to the ambiguous nature of some class names benefited from the addition of definitions. Lastly, classes such as the "airplane" or "elephant" class benefited due to size, color, or shape descriptions present in the definitions which provided a more accurate visual-semantic embedding. While many class APs decreased with the addition of definitions, we believe that adding definitions to inference prompts remains a simple and effective way to provide better visual-semantic embeddings for ambigious class names. Furthermore, we believe that the addition of definitions to prompts do not violate ZSD benchmarking rules given that other methods are able to obtain proper word embeddings by simply hand selecting the correct WordNet synset, a method that is not available to CLIP. When applying a ZSD method to real world problems, users will likely have at least one example of their target class to test inference, and therefore can tune their prompts manually or use vision-language prompt engineering methods~\cite{DBLP:journals/corr/abs-2109-01134}, though we avoided doing so in our paper as ZSD methods cannot prompt tune in this manner. 

\begin{table*}[t]
  \caption{Ablation study class-wise AP for unseen classes of COCO 65/15 ZSD split\cite{Rahman_aaai_2020}. ZSD-YOLOv5x denotes not using definitions, ZSD-YOLOv5x + definition denotes using definitions. Note that the mean for Gain Percentage is the actual mean of the gain percentages, not the percent increase in mAP. }

  \begin{center}\setlength\tabcolsep{2pt}
  \resizebox{\linewidth}{!}{
    \begin{tabular}{c|c||c|c|c|c|c|c|c|c|c|c|c|c|c|c|c|c|c}
    \toprule[0.1em]
    
\rotatebox{90}{ }&\rotatebox{90}{ \textbf{mean}}&\rotatebox{90}{tie}&\rotatebox{90}{airplane}&\rotatebox{90}{elephant}&\rotatebox{90}{dog}&\rotatebox{90}{knife}&\rotatebox{90}{sink}&\rotatebox{90}{bus}&\rotatebox{90}{snowboard}&\rotatebox{90}{cat}&\rotatebox{90}{cake}&\rotatebox{90}{cow}&\rotatebox{90}{cup}&\rotatebox{90}{scissors}&\rotatebox{90}{couch}&\rotatebox{90}{keyboard}&\rotatebox{90}{skateboard}&\rotatebox{90}{umbrella}\\
\toprule[0.1em]

ZSD-YOLOv5x & 13.0 & 0.0 & 10.5 & 18.8 & 11.3 & 1.5 & 3.3 & 51.6 & 21.6 & 5.3 & 5.7 & 34.9 & 13.3 & 0.1 & 31.9 & 8.2 & 0.6 & 2.6\\
ZSD-YOLOv5x + definition & 13.4 & 0.0 & 19.0 & 29.6 & 16.7 & 2.1 & 4.5 & 62.9 & 24.1 & 5.8 & 4.9 & 27.9 & 10.4 & 0.1 & 15.5 & 3.7 & 0.2 & 0.1\\
Gain & +0.4 & 0.0 & +8.6 & +10.8 & +5.4 & +0.6 & +1.1 & +11.3 & +2.5 & +0.5 & -0.8 & -7.0 & -3.0 & -0.0 & -16.4 & -4.5 & -0.5 & -2.5\\
Gain Percentage & +47.6\% & +865.3\% & +81.9\% & +57.5\% & +48.2\% & +42.4\% & +34.3\% & +22.0\% & +11.6\% & +9.7\% & -13.2\% & -20.1\% & -22.2\% & -33.2\% & -51.3\% & -54.6\% & -73.0\% & -95.4\\

\bottomrule[0.1em]
    \end{tabular}}
  \end{center}
  \label{tab:COCO_48_17_def_ablation}
\end{table*}

\begin{table*}[t]
  \caption{Ablation study class-wise AP for unseen classes of COCO 65/15 ZSD split\cite{Rahman_aaai_2020}. ZSD-YOLOv5x denotes not using definitions, ZSD-YOLOv5x + definition denotes using definitions. Note that the mean for Gain Percentage is the actual mean of the gain percentages, not the percent increase in mAP.}

  \begin{center}\setlength\tabcolsep{2pt}
  \resizebox{\linewidth}{!}{
    \begin{tabular}{c|c||c|c|c|c|c|c|c|c|c|c|c|c|c|c|c}
    \toprule[0.1em]
    
\rotatebox{90}{ }&\rotatebox{90}{ \textbf{mean}}&\rotatebox{90}{bear}&\rotatebox{90}{cat}&\rotatebox{90}{airplane}&\rotatebox{90}{parking meter}&\rotatebox{90}{mouse}&\rotatebox{90}{toaster}&\rotatebox{90}{train}&\rotatebox{90}{hair drier}&\rotatebox{90}{hot dog}&\rotatebox{90}{frisbee}&\rotatebox{90}{snowboard}&\rotatebox{90}{fork}&\rotatebox{90}{suitcase}&\rotatebox{90}{toilet}&\rotatebox{90}{sandwich} \\
\toprule[0.1em]

ZSD-YOLOv5x & 14.9 & 19.6 & 16.5 & 14.0 & 1.7 & 2.9 & 1.1 & 28.4 & 0.5 & 7.4 & 28.0 & 39.2 & 16.4 & 11.2 & 17.6 & 18.9 \\
ZSD-YOLOv5x + definition & 18.3 & 54.0 & 36.3 & 18.3 & 2.0 & 3.3 & 1.2 & 32.2 & 0.6 & 7.9 & 29.4 & 39.5 & 15.1 & 10.1 & 12.9 & 11.8 \\
Gain & +3.4 & +34.4 & +19.8 & +4.3 & +0.3 & +0.4 & +0.1 & +3.7 & +0.1 & +0.5 & +1.4 & +0.4 & -1.3 & -1.0 & -4.7 & -7.0 \\
Gain Percentage & +21.9\% & +175.1\% & +120.0\% & +30.4\% & +20.0\% & +14.3\% & +13.5\% & +13.1\% & +10.1\% & +6.7\% & +5.0\% & +0.9\% & -7.9\% & -9.0\% & -26.7\% & -37.3 \\

\bottomrule[0.1em]
    \end{tabular}}
  \end{center}
  \label{tab:COCO_65_15_def_ablation}
\end{table*}

\section{Validation Set Details}
\label{Validation Set Details}
\begin{figure*}[t]
\centering
   \includegraphics[width=0.95\linewidth]{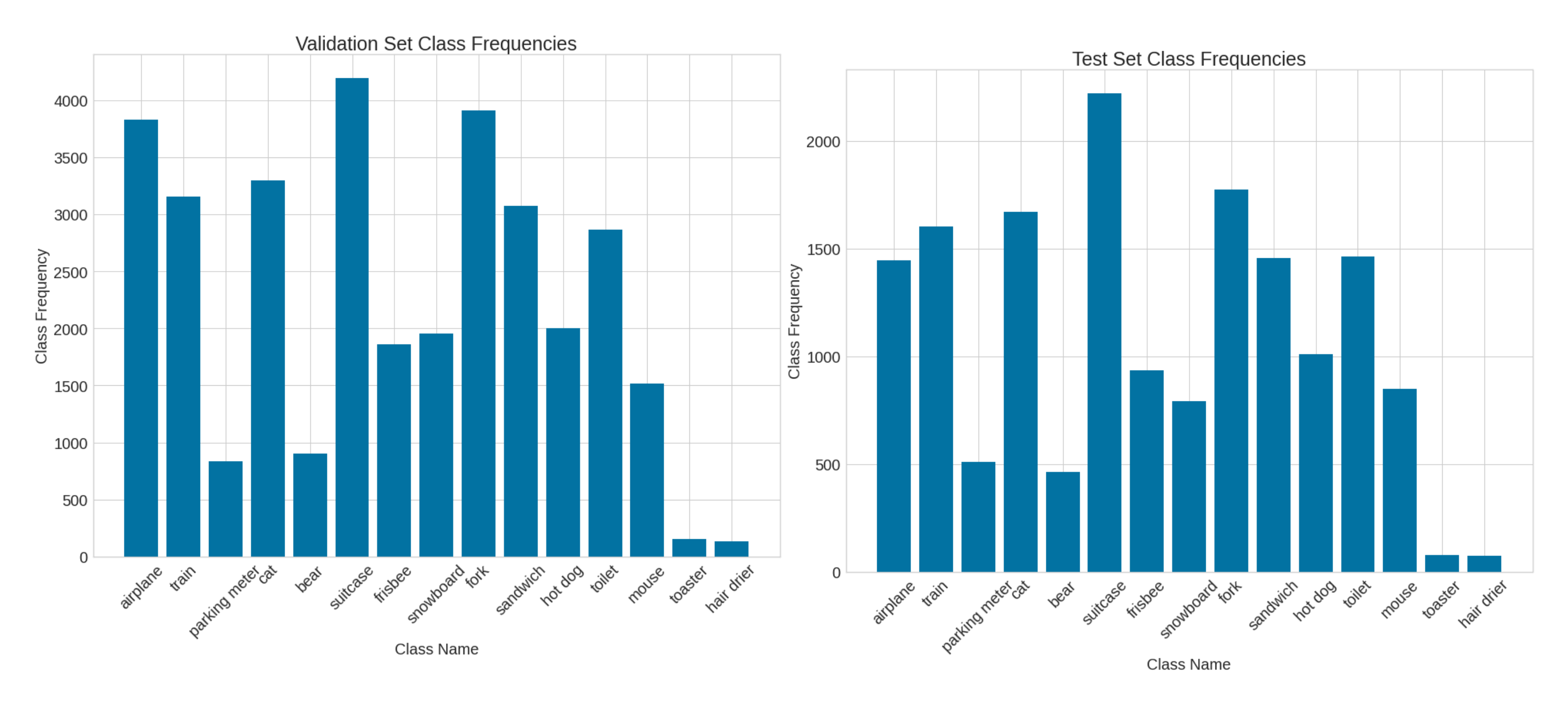}
   \caption{Histogram comparison of class frequencies between proposed validation set and typical testing set of COCO 65/15 ZSD class split. The validation set is created by taking the portion of the COCO 2014 training split that is not used in 65/15 ZSD training or testing. 
   }
\label{supp_fig_65_15_val_test_hist_extra_2}
\end{figure*}

While all other ZSD benchmark datasets used in our main paper have been developed in prior works~\cite{Zheng_zsis_cvpr2021,Bansal_eccv_2018,Rahman_aaai_2020,Rahman_accv_2018}, we provide the details of our proposed validation set for future use. The validation set is created by taking the portion of the COCO 2014 training split that is not used in 65/15 ZSD training or testing. This validation set contains 20,483 images with 33,690 unseen box annotations. Our validation set has the same unseen classes as the COCO 65/15 testing set. We compare the class distributions in Figure~\ref{supp_fig_65_15_val_test_hist_extra_2}, and find that the two distributions are relatively similar with a mean-normalized chi-square distance of approximately 0.206.

We did not create a validation set for the 48/17 COCO split and other benchmarks because our developed methods were transferred from the 65/15 COCO split. Our class-wise AP results for the COCO 65/15 split and 48/17 split are displayed in Table \ref{tab:COCO 65_15_class_AP} and Table \ref{tab:COCO 48_17_class_AP} respectively.

\section{Qualitative Results}
\label{Qualitative Results}
\begin{figure*}[t]
\centering
   \includegraphics[width=0.95\linewidth]{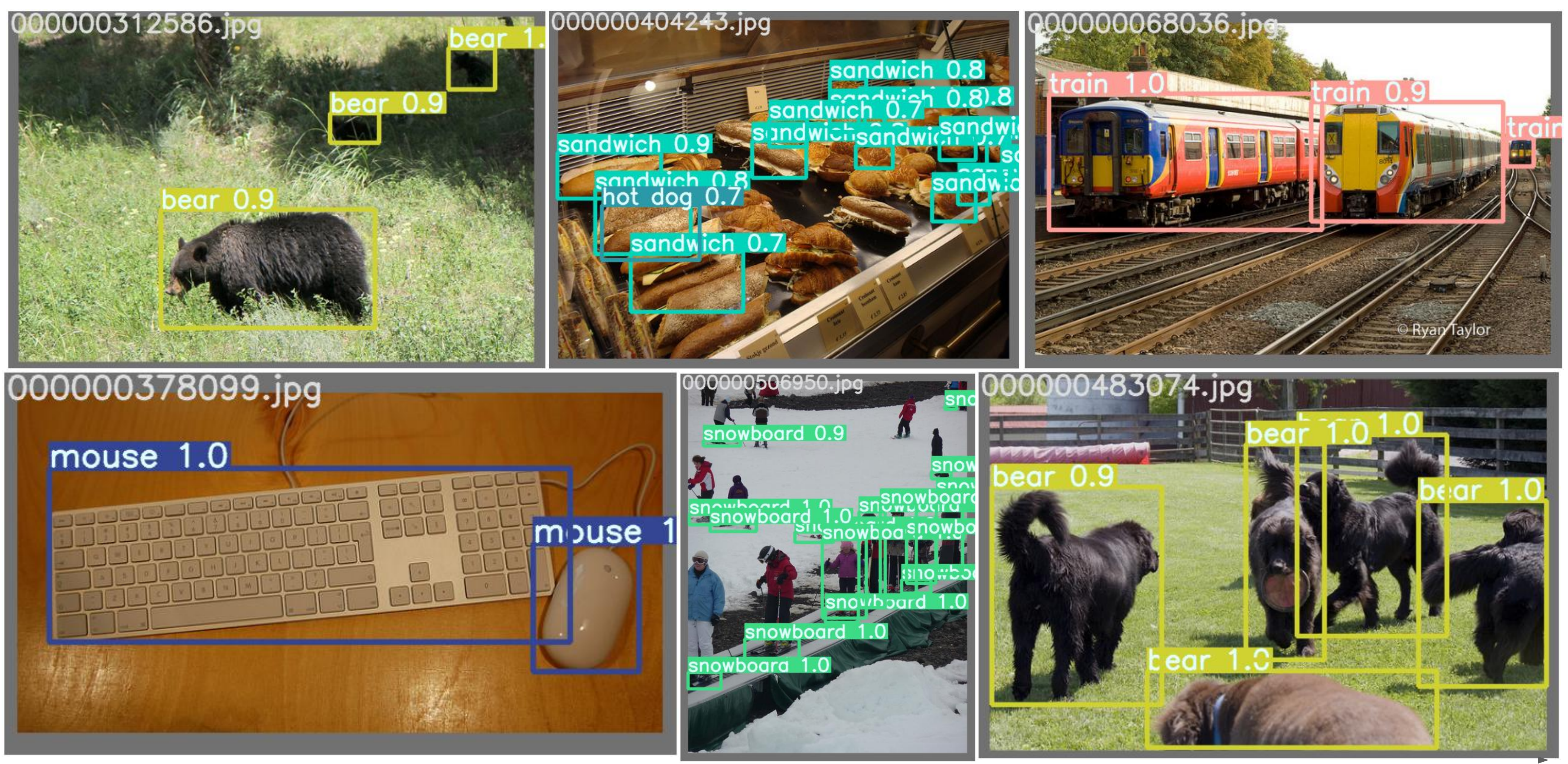}
   \caption{
   Selected examples of our results for zero shot detection produced by our ZSD-YOLOv5x model on the COCO 65/15 testing set. Well detected images are in the top row while poorly detected images are displayed in the bottom row. Best view in color.
   }
\label{fig:qual_results}
\end{figure*}

We display a couple examples of our zero-shot detector on the COCO 65/15 testing dataset split in Figure~\ref{fig:qual_results}. We find our proposed model is able to detect diverse unseen classes under various situations with well refined boxes. The top left image shows our model's ability accurately detect the unseen class of "bear" despite two instances being obscured by grass and shadow. The top middle image reveals our model's ability to perform dense zero shot detection as it detects most of the sandwiches while both filtering our similar objects that are not labeled as sandwiches and correctly classifying the hot dog. Lastly the top right image show our model's ability to differentiate between multiple similar and overlapping objects of the "train" class against a noisy background. When examining the mistakes of our model displayed in the bottom row, we mainly see an issue with false positives of seen classes being recognized as unseen classes. In the images shown in the bottom row, we see "keyboard" being detected as "mouse," "person" being detected as "snowboard," and "dog" being detected as "bear." We believe that our softmax based class prediction is a cause for this issue since even when a semantic output is dissimilar from a reference class embedding, if it is relatively close compared to other reference embeddings, it may be assigned a high confidence. To address this issue we experimented with a sigmoid based activation where the class confidences would not be dependent on similarity to other class embeddings. This method did not perform well in training because when applying a "temperature" factor similar to that found in temperatured softmax to sigmoid, a majority of gradients vanish due to extremely low gradients on the left and right sides of the activation. Further research will be needed to work to better filter out objects that are not present in the given embedding set from being detected.

\section{Further Implementation Details}
\label{Further Implementation Details}
\paragraph{\textbf{Hyperparameter search.}}
After conducting pretraining as described in our main paper, we fit detection training parameters for our ZSD-YOLOv5x model using mAP on our proposed COCO 65/15 validation set as the fitness parameter. Our final hyperparameter search values are listed in Table \ref{tab:hyperparameters}. Across all dataset splits and evaluation benchmarks, the only change made to hyperparameters is a heuristic increase in temperature value for the Visual Genome ZSD benchmark~\cite{Bansal_eccv_2018} due to the greater number of classes present.

\paragraph{\textbf{Post-processing.}}
For our proposed post-processing function, we use the following parameters. NMS-IOU-threshold = 0.4, initial-confidence-threshold = 0.001, second-confidence-threshold\footnote{Second confidence cutoff applied to confidence values that are equal to the product of objectness score and max class similarity} = 0.1, max-detection = 15. Across datasets and benchmarks certain post-processing parameters change heuristically. For example in GZSD, max-detections increases to 45, and in Recall@100 evaluations max-detections increases to 100. Confidence values also change depending on the benchmark. Full details and paramters are shown in our code implementation attached.

\paragraph{\textbf{Self-labeling.}}
For self-labeling experiments we limit self-label boxes to only those which are greater than 25 pixels in both width and height. An initial objectness cutoff of 0.3 is applied followed by IOU with 0.2 threshold to maximize unique boxes. Self-labeling is performed only for training sets which will not be released in the initial supplementary materials due to file size limitations. Still, all weights are trained via self-labeling and are available in the initial attached release.

\section{Visual Genome Ablation}
\label{Visual Genome Ablation}
To further verify the efficacy of our proposed post-processing function we perform an additional ablation experiment on the Visual Genome ZSD benchmark~\cite{Bansal_eccv_2018} to examine how our post-processing affects both Recall@100 and mAP. We present our results in Table~\ref{table:vg_zsd_post}. We find similar results to those of our COCO ablation experiments with our proposed post-processing performing better under all benchmarks except for Recall@100 IOU=0.6. We believe this is largely because our method factors objectness score less and therefore lower objectness boxes may be favored over higher objectness boxes that typically match with ground truth box more closely in NMS. To address this issue, we also create another post-processing function denoted as ZSD-POST+ in the table. In this operation, we factor objectness into the NMS value by multiplying the maximum similarity value of each class before NMS then only using maximum similarity class values for final confidence. In comparison, our usual proposed post-processing uses only maximum class values for NMS and final confidence scores. For the Visual Genome benchmark we find that this operation provides the best performance under all listed recall benchmarks. While we also experimented with this function on the original COCO benchmarks, we did not find improvements. For example on the COCO 65/15 benchmark~\cite{Rahman_aaai_2020} we find that this function lowers mAP from 18.3 to 17.6. Therefore, in our main paper we only use the original post processing function, ZSD-POST, as we must transfer all methods across benchmarks with only minor parameter changes.

\begin{table}[t]
\caption{ZSD training settings for COCO dataset splits. Augmentations are shown in the second half of the table and their likelihoods are indicated as a ratio between 0 and 1.}
\label{tab:hyperparameters}
\scriptsize
\begin{tabular}{c|c}
config & value \\
\hline
optimizer & SGD \\
base learning rate & 0.00282 \\
weight decay & 0.00038 \\
optimizer momentum & 0.85403 \\
batch size & 20 \\
learning rate schedule & cosine decay \\
positive iou threshold & 0.14671 \\
box loss weight & 0.03 \\
class loss weight & 0.04688 \\
objectness loss weight & 2.67895 \\
temperature & 3.91 \\
\hline
image rotation (+/- deg) & 0.373 \\
image flip left-right (probability)  & 0.5 \\
image flip up-down (probability)  & 0.00856 \\
image HSV-Hue augmentation (fraction) & 0.0168 \\
image HSV-Saturation augmentation (fraction) & 0.7876 \\
image HSV-Value augmentation (fraction) & 0.45518 \\
image mosaic augmentation (probability) & 0.94109 \\
image scale (+/- gain)  & 0.64818 \\
image shear (+/- deg)  & 0.602 \\
image translation (+/- fraction)  & 0.16587 \\
\end{tabular}
\label{tab:impl_mae_pretrain}
\end{table}

\begin{table}[!t]
\caption{Comparison of effect of proposed post-processing on the Visual Genome Benchmark. YOLO-post stands for the typical baseline post-processing, ZSD-post stands for our proposed post-processing, and ZSD-post+ stands for our proposed post-processing that uses objectness in NMS but not in the final confidence prediction.}
\label{table:vg_zsd_post}
\begin{center}
\resizebox{\linewidth}{!}{
\begin{tabular}{c|c|c|c|c}
\toprule
\multirow{2}{*}{Method} & \multicolumn{3}{c}{Recall@100} & mAP \\
\cmidrule(lr){2-4} \cmidrule(lr){5-5}
~ & 0.4 & 0.5 & 0.6 & 0.5 \\
\midrule
ZSD-YOLOv5x + YOLO-POST & 24.6 & 22.0 & \textbf{18.5} & 1.64 \\
ZSD-YOLOv5x + ZSD-POST & 27.8 & 22.3 & 15.1 & \textbf{1.96} \\
ZSD-YOLOv5x + ZSD-POST+ & \textbf{28.6} & \textbf{24.4} & 18.4 & 1.73 \\
\bottomrule
\end{tabular}}
\end{center}
\end{table}

\begin{table*}[t]
  \caption{ZSD class AP for unseen classes of COCO 65/15 ZSD split\cite{Rahman_aaai_2020} for 4 given YOLO model variants. Note that all methods use both our ZSD tailored post-processing function and self-labeling augmentation. Note that scaling improvements are relatively consistent across class APs showing that typical detector scaling rules can transfer well to our ZSD method.}

  \begin{center}\setlength\tabcolsep{2pt}
  \resizebox{\linewidth}{!}{
    \begin{tabular}{c|c||c|c|c|c|c|c|c|c|c|c|c|c|c|c|c}
    \toprule[0.1em]
    
\rotatebox{90}{ }&\rotatebox{90}{ \textbf{mean}}&\rotatebox{90}{airplane}&\rotatebox{90}{train}&\rotatebox{90}{parking meter}&\rotatebox{90}{cat}&\rotatebox{90}{bear}&\rotatebox{90}{suitcase}&\rotatebox{90}{frisbee}&\rotatebox{90}{snowboard}&\rotatebox{90}{fork}&\rotatebox{90}{sandwich}&\rotatebox{90}{hot dog}&\rotatebox{90}{toilet}&\rotatebox{90}{mouse}&\rotatebox{90}{toaster}&\rotatebox{90}{hair drier}\\
\toprule[0.1em]
ZSD-YOLOv5x & 18.3 & 18.3 & 32.2 & 2.0 & 36.3 & 54.0 & 10.1 & 29.4 & 39.5 & 15.1 & 11.8 & 7.9 & 12.9 & 3.3 & 1.2 & 0.6\\
ZSD-YOLOv5m & 17.4 & 23.2 & 29.3 & 6.0 & 32.8 & 50.9 & 6.8 & 27.3 & 36.8 & 13.4 & 10.5 & 7.3 & 10.1 & 5.1 & 0.9 & 0.8\\
ZSD-YOLOv5s & 14.0 & 15.7 & 24.6 & 6.2 & 28.9 & 46.3 & 5.9 & 21.9 & 29.1 & 8.6 & 7.2 & 3.9 & 7.1 & 4.8 & 0.5 & 0.0\\
ZSD-YOLOv3 & 16.8 & 20.3 & 28.3 & 4.4 & 30.7 & 50.9 & 9.4 & 26.6 & 40.9 & 10.1 & 9.1 & 6.8 & 10.7 & 2.9 & 1.0 & 0.5\\

\bottomrule[0.1em]
    \end{tabular}}
  \end{center}
  \label{tab:COCO 65_15_class_AP}
\end{table*}

\begin{table*}[t]
  \caption{ZSD class AP for unseen classes of COCO 48/17 ZSD split\cite{Bansal_eccv_2018} for 4 given YOLO model variants. Note that all methods use both our ZSD tailored post-processing function and self-labeling augmentation. Note that scaling improvements are relatively consistent across class APs showing that typical detector scaling rules can transfer well to our ZSD method.} 

  \begin{center}\setlength\tabcolsep{2pt}
  \resizebox{\linewidth}{!}{
    \begin{tabular}{c|c||c|c|c|c|c|c|c|c|c|c|c|c|c|c|c|c|c}
    \toprule[0.1em]
    
\rotatebox{90}{ }&\rotatebox{90}{ \textbf{mean}}&\rotatebox{90}{airplane}&\rotatebox{90}{bus}&\rotatebox{90}{cat}&\rotatebox{90}{dog}&\rotatebox{90}{cow}&\rotatebox{90}{elephant}&\rotatebox{90}{umbrella}&\rotatebox{90}{tie}&\rotatebox{90}{snowboard}&\rotatebox{90}{skateboard}&\rotatebox{90}{cup}&\rotatebox{90}{knife}&\rotatebox{90}{cake}&\rotatebox{90}{couch}&\rotatebox{90}{keyboard}&\rotatebox{90}{sink}&\rotatebox{90}{scissors}\\
\toprule[0.1em]
ZSD-YOLOv5x & 13.4 & 19.0 & 62.9 & 5.8 & 16.7 & 27.9 & 29.6 & 0.1 & 0.0 & 24.1 & 0.2 & 10.4 & 2.1 & 4.9 & 15.5 & 3.7 & 4.5 & 0.1\\
ZSD-YOLOv5m & 11.4 & 13.0 & 57.4 & 4.3 & 20.4 & 19.7 & 22.2 & 0.1 & 0.0 & 19.4 & 0.4 & 8.1 & 2.1 & 6.5 & 11.9 & 3.7 & 4.4 & 0.2\\
ZSD-YOLOv5s & 10.0 & 12.1 & 55.3 & 4.6 & 14.9 & 16.7 & 17.6 & 0.1 & 0.0 & 17.8 & 0.3 & 7.3 & 3.1 & 4.4 & 10.5 & 1.6 & 3.6 & 0.1\\
ZSD-YOLOv3 & 11.2 & 9.3 & 58.0 & 4.7 & 16.3 & 23.3 & 23.4 & 0.2 & 0.0 & 20.3 & 0.2 & 7.7 & 2.3 & 4.0 & 12.4 & 2.4 & 5.8 & 0.1\\

\bottomrule[0.1em]
    \end{tabular}}
  \end{center}
  \label{tab:COCO 48_17_class_AP}
\end{table*}

\end{document}